%% file: main.tex
\documentclass[10pt,twocolumn,letterpaper]{article}

\usepackage[pagenumbers]{cvpr}


\definecolor{cvprblue}{rgb}{0.21,0.49,0.74}
\usepackage[pagebackref,breaklinks,colorlinks,allcolors=cvprblue]{hyperref}
\usepackage[accsupp]{axessibility}  

\def\paperID{49} 
\def\confName{CVPR}
\def\confYear{2025}

\title{Ice Hockey Puck Localization Using Contextual Cues}

\author{
Liam Salass, Jerrin Bright, Amir Nazemi, Yuhao Chen, John Zelek, David Clausi\\
{\tt\small \{liam.salass, jerrin.bright, amir.nazemi, yuhao.chen1, jzelek, dclausi\}@uwaterloo.ca}\\
University of Waterloo, Waterloo, ON, Canada, N2L 3G1
}

\begin{document}
\maketitle
\input{sec/0_abstract}    
\input{sec/1_intro}
\input{sec/2_related_works}
\input{sec/3_methodology}

\input{sec/4_experiments}
\input{sec/5_conclusions}
\input{sec/acknowledgments}

{
    \small
    \bibliographystyle{ieeenat_fullname}
    \bibliography{main}
}


\end{document}

%% file: sec/0_abstract.tex
\begin{abstract}
Puck detection in ice hockey broadcast videos poses significant challenges due to the puck's small size, frequent occlusions, motion blur, broadcast artifacts, and scale inconsistencies due to varying camera zoom and broadcast camera viewpoints. Prior works focus on appearance-based or motion-based cues of the puck without explicitly modelling the cues derived from player behaviour. Players consistently turn their bodies and direct their gaze toward the puck. Motivated by this strong contextual cue, we propose Puck Localization Using Contextual Cues (PLUCC), a novel approach for scale-aware and context-driven single-frame puck detections. PLUCC consists of three components: (a) a contextual encoder, which utilizes player orientations and positioning as helpful priors; (b) a feature pyramid encoder, which extracts multiscale features from the dual encoders; and (c) a gating decoder that combines latent features with a channel gating mechanism. For evaluation, in addition to standard average precision, we propose Rink Space Localization Error (RSLE), a scale-invariant homography-based metric for removing perspective bias from rink space evaluation. The experimental results of PLUCC on the PuckDataset dataset demonstrated state-of-the-art detection performance, surpassing previous baseline methods by an average precision improvement of 12.2\% and RSLE average precision of 25\%. Our research demonstrates the critical role of contextual understanding in improving puck detection performance, with broad implications for automated sports analysis.
\end{abstract}

%% file: sec/1_intro.tex
\section{Introduction}


Computer vision has been widely used for ice hockey analytics \cite{track1, track2, pose1, pose2, jersey1, jersey2, event1, vats2021puck, tian2018group}. The applications have expanded from player detection/ tracking \cite{track1, track2}, pose estimation \cite{pose1, pose2}, and jersey number recognition \cite{jersey1, jersey2} to more advanced tasks, including gameplay strategy analysis \cite{event1, vats2021puck, tian2018group}, and puck possession estimation \cite{possession1, possession2}. However, a common underlying factor in most of these tasks is the understanding of the precise location of the puck, which provides crucial insight into game dynamics.


\begin{figure}[t]
    \centering
    \begin{subfigure}{0.48\columnwidth}
        \centering
        \includegraphics[width=\linewidth]{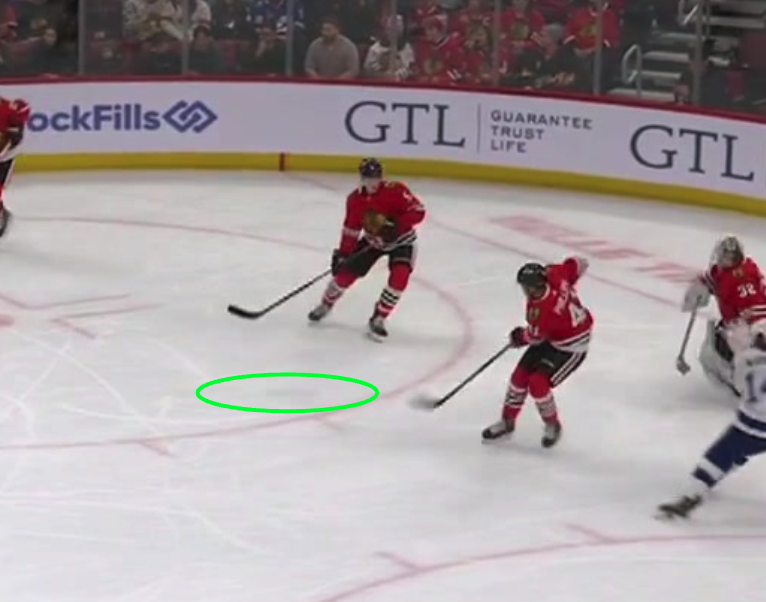}
        \caption{Blur.}
        \label{fig:dif_blur}
    \end{subfigure}
    \hfill
    \begin{subfigure}{0.48\columnwidth}
        \centering
        \includegraphics[width=\linewidth]{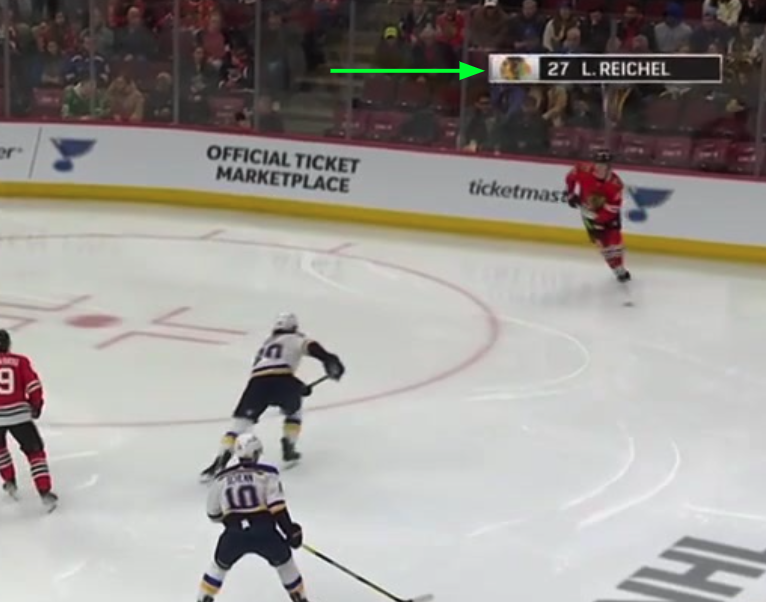}
        \caption{Artifact.}
        \label{fig:dif_artifact}
    \end{subfigure}
    
    \vspace{0.3em} 

    \begin{subfigure}{0.48\columnwidth}
        \centering
        \includegraphics[width=\linewidth]{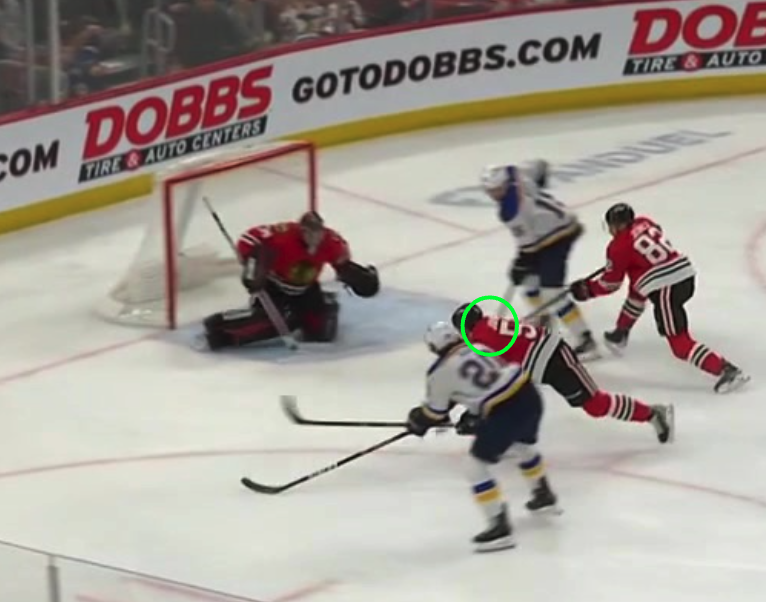}
        \caption{Occluded.}
        \label{fig:dif_occluded}
    \end{subfigure}
    \hfill
    \begin{subfigure}{0.48\columnwidth}
        \centering
        \includegraphics[width=\linewidth]{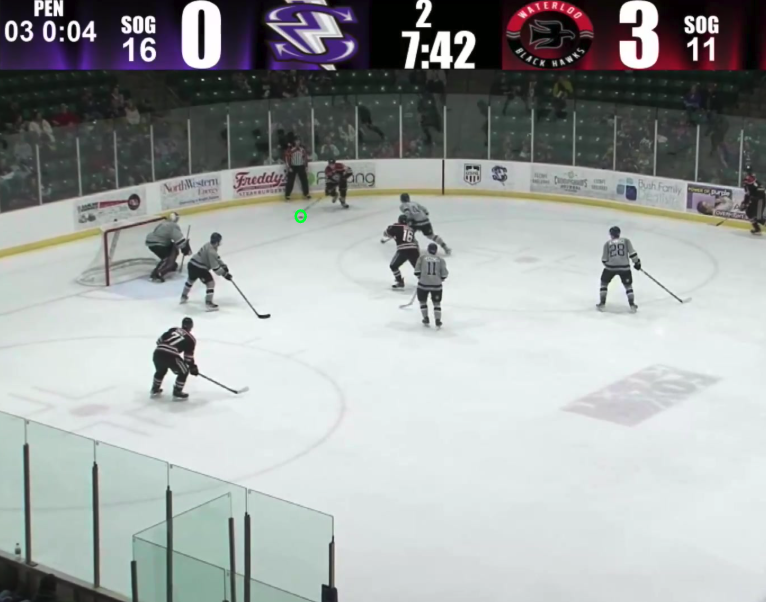}
        \caption{Small.}
        \label{fig:dif_small}
    \end{subfigure}

     \vspace{0.3em} 

    \begin{subfigure}{0.48\columnwidth}
        \centering
        \includegraphics[width=\linewidth]{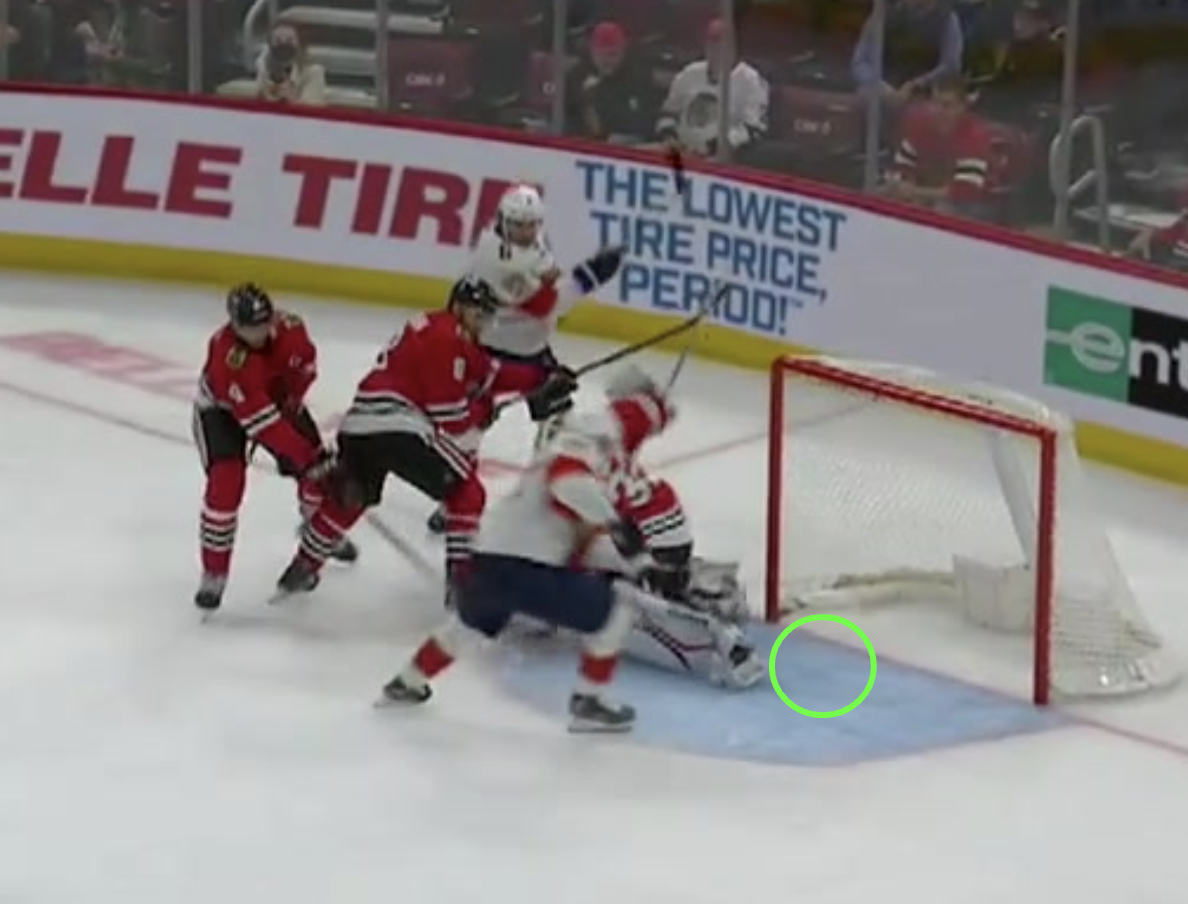}
        \caption{Goal-line puck.}
        \label{fig:dif_goal}
    \end{subfigure}
    \hfill
    \begin{subfigure}{0.48\columnwidth}
        \centering
        \includegraphics[width=\linewidth]{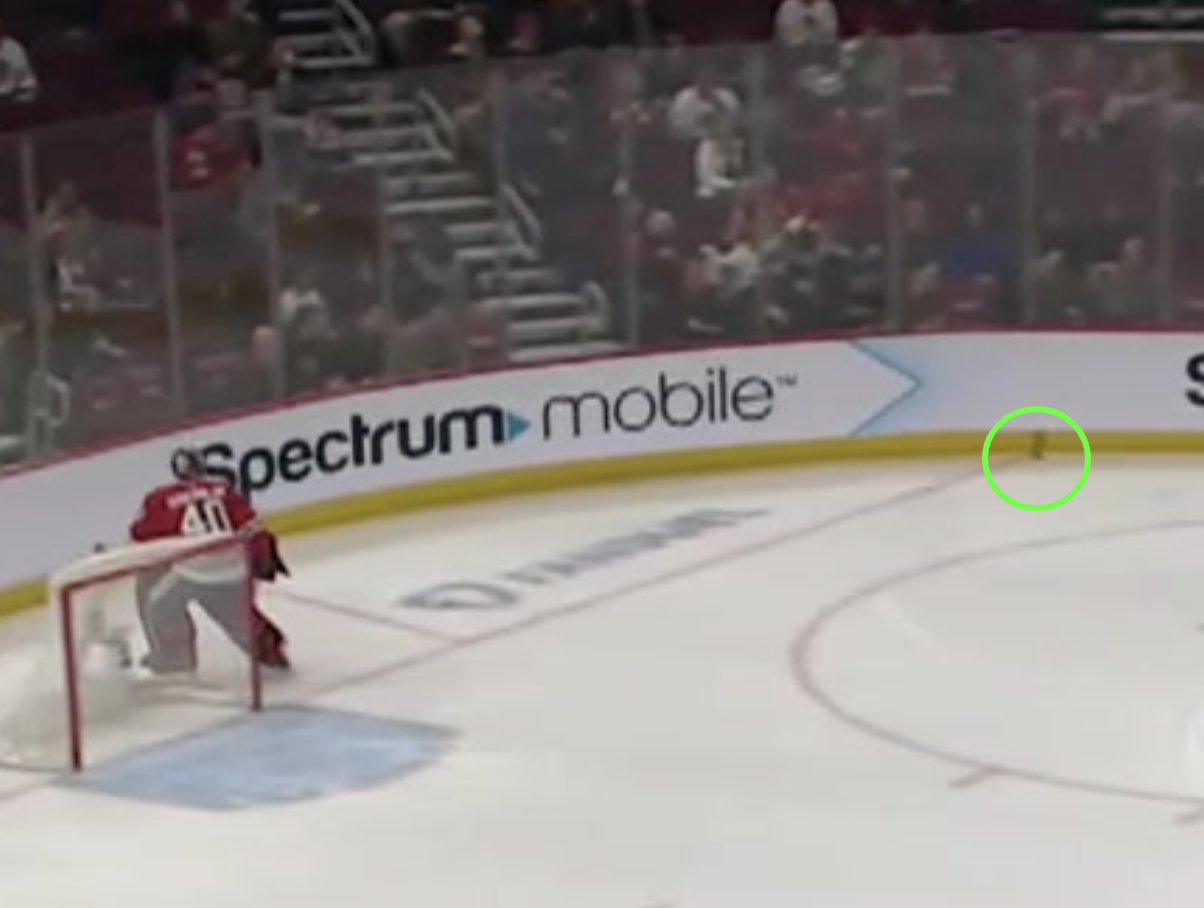}
        \caption{Yellow-line puck.}
        \label{fig:dif_yellow}
    \end{subfigure}
\caption{Test set examples of challenges faced in automated puck detection from ice hockey broadcast video: (a) Motion blur causing deformation; (b) Artifacts introduced by broadcast overlays; (c) Occlusions from player bodies obstructing camera views; (d) Small puck size, occupying only approximately 0.005\% of the frame pixels; (e) Goal-line puck where the puck contour is harder to visualize due to the change in contrast; (d) the puck on the yellow-line of rink borders being difficult to distinguish do to similar colours.}
\label{fig:difficult}
\vspace{-4mm}
\end{figure}

Given its fundamental importance in ice hockey analytics, accurate puck detection from broadcast video is essential~\cite{trackingimportant}, developing systems that assist coaches~\cite{kinexon2025}, and a general understanding of the ice hockey game. Although ball detection and tracking has been extensively studied in different sports such as soccer~\cite{4607578, Beetz2009ASPOGAMOAS, probtracksoccer, tvtracksoceer, leo_marco, REN2009633,VicenteMartnez2023AdaptationOY, balltracksoccer}, volleyball~\cite{4217275, volleyballtrajectory, chengvolley, chengnumber2}, and basketball~\cite{chakbasket, Chakraborty2012RealtimePE, chakdontmiss}, there is limited research on puck tracking in ice hockey games~\cite{vats2021puck, li2023ice, sarkhoosh_ai-based_2024, pidaparthy2019keep, yang2021puck}. 

Prior works in automated puck detection have addressed challenges such as occlusions, motion blur, visual perspective distortion, and more (highlighted in Figure \ref{fig:difficult}) by relying on manual thresholding, temporal cues, or low-level contextual features \cite{li2023ice, vats2021puck, yang2021puck, pidaparthy2019keep}. Many methods filter out incorrect detections by leveraging temporal consistency \cite{yang2021puck, vats2021puck, li2023ice} or employ course player masks, positions, and flow maps to add context \cite{pidaparthy2019keep, vats2021puck}. However, these approaches do not fully capture the explicit cues between player poses and positions with puck motion—especially under significant viewpoint variations and broadcast-induced zoom distortions. 

In contrast, our proposed method, PLUCC, produces robust detections, even under stick occlusions, leveraging RGB-based player segmentations that capture pose and position information as helpful priors. This rich contextual cue enables our model to accurately localize the puck on a per-frame basis without any reliance on temporal detections. The single-frame processing not only simplifies the detection pipeline but also avoids the error propagation typically observed in multi-frame temporal methods \cite{lu2020content}.   

Our approach amalgamates a context encoder within a feature pyramid network, enabling a multi-scale fusion of spatial and contextual information. This design improves detection accuracy in challenging scenarios such as occlusions, motion blur, yellow-line pucks, and goal-line pucks. Our experiments on the VIP-PuckDataset highlight significant improvements, achieving a 25\% reduction in Rink-Space Localization Error (RSLE), a novel metric for comparing puck detection accuracies within the coordinate space of the rink, and a 12\% boost in average precision compared to baseline models. The combination of RGB player segmentations and single-frame processing differentiates PLUCC from prior methods. 

Our main contributions can be summarized as follows:
\begin{itemize}
    \item We leverage player RGB masks as explicit priors, guiding our detection network to make more accurate detections, resulting in a 7.4\% increase in average precision at a fine-grained threshold.
    \item Propose PLUCC, a multi-scale pyramid network incorporating context features with gated channel fusions, achieving a 12.6\% increase in average precision under a stringent detection threshold.
    \item Propose a new metric for evaluating puck detection accuracy in rink-space coordinates, providing an empirical unit of measure for comparison and is invariant to the perspective distortion of broadcast views of the puck. 
\end{itemize}

%% file: sec/2_related_works.tex
\section{Related Works}


\subsection{Puck Detection}
There are five main methods for regressing the puck's location \cite{yang2021puck, pidaparthy2019keep, li2023ice, vats2021puck, sarkhoosh_ai-based_2024}, often leveraging a combination of deep learning, temporal tracking, and/or classical detection techniques, and contextual cues.

Yang \cite{yang2021puck} highlights the shortcomings of the dated YOLOv3 \cite{yolov3} and Mask-RCNN \cite{maskrcnn} models on puck detection, notes their high false positive rates due to their model architecture being built around multiple object detection, and proposes a novel deep-learning architecture leveraging multi-headed learning and temporal features. The method requires a sequence of nine frames to regress a single detection, differentiating itself from the PLUCC model. 

Pidaparthy et al.'s \cite{pidaparthy2019keep} system locates the location of play by regressing the puck location using a deep neural network. Their method leverages estimated player locations, optical flow, and the regressed puck location, enabling their system to move a camera to focus on a region of play. Pidaparthy et al. \cite{pidaparthy2019keep} note the need for more diverse and more extensive datasets and mention that player orientation and actions can be used to refine these systems further.  

Li et al. \cite{li2023ice} propose a tracking method that classifies puck motion into controlled and free-moving states. Their approach leverages image matching the puck for instances where it is visible and motion estimation during occlusions. However, Li et al. note the technique is prone to false positives because artifacts—such as the players' skates or noise in the background—can be misinterpreted as the puck.

Vats et al.'s \cite{vats2021puck} PuckNet leverages a 3D convolutional neural network to regress a soft heatmap from short video clips representing the puck's general location. This network leverages temporal context to mitigate occlusion issues; however, it struggles with the puck’s inherent small size and rapid motion, which can lead to mislocalizations. Furthermore, the model relies on the contextual cue that the puck is typically located in regions with high player density—an assumption that can result in localization errors when the puck ventures into less congested areas. Furthermore, their model was only trained on the relative location of the puck and could not directly regress its precise location. 
    
Most recently, Sarkhoosh et al. \cite{sarkhoosh_ai-based_2024} propose an AI-based video cropping pipeline to tailor hockey video content for social media. Their method uses fine-tuned detection models such as Faster-RCNN \cite{ren_faster_2016} and YOLOv8 \cite{yolov8} to find regions of interest in the match by regressing the puck's location, noting that YOLOv8 X-Large outperformed all other models. 



\subsection{Object Detection}

Wei et al. \cite{survey_small_obj_wei} survey small object detection techniques; their findings offer insights on applicability to ice hockey puck detection, specifically in enhancing input feature resolution, scale-aware training, and incorporating contextual information. They explore contextual information that can be integrated through mechanisms such as attention or squeeze-and-excitation, helping models focus on relevant areas. This idea motivated PLUCC's context encoding. 

In their baseline for sports ball detection algorithms, Tarashima et al. \cite{tarashima_widely_2023} leverage heatmap labels and high-resolution feature extraction networks, achieving high detection accuracies in multiple sports with balls of varying sizes. They argue that heatmap labels are necessary for their ease of integration with tracking methods and superior accuracy, thus inspiring the PLUCC label representation. Lastly, the high-resolution feature extraction model coincides with Wei et al.'s findings on the importance of preservation of high-resolution image inputs. 

\subsection{Contextually Constraint Detection}
Contextual constraints in detection algorithms are standard performance boosters in detection tasks where semantic correlations between spatial positions and object properties refine detection probabilities \cite{high_freq_context, transformer_context, DBLP:journals/corr/GidarisK15}. Shi et al. \cite{high_freq_context} enrich a feature pyramid network for object detection using intermediary lateral modules that provide latent features spatial and high-frequency correlations, demonstrating improved performance for baseline models including Faster-RCNN \cite{ren_faster_2016}. This lateral context fusing method inspired the static gating fusion in the PLUCC architecture.  

%% file: sec/3_methodology.tex
\section{Methodology}
This section outlines the PLUCC framework for robust puck detection. The proposed model architecture consists of three main components: (a) the feature pyramid encoder, (b) the context encoder, and (c) the gated decoder. Lastly, our method’s performance depends on the objective function, which leverages Kullback-Leibler divergence loss with Gaussian labels.


\subsection{Model Architecture}
\begin{figure*}[t]
    \centering
    \includegraphics[width=0.9\textwidth]{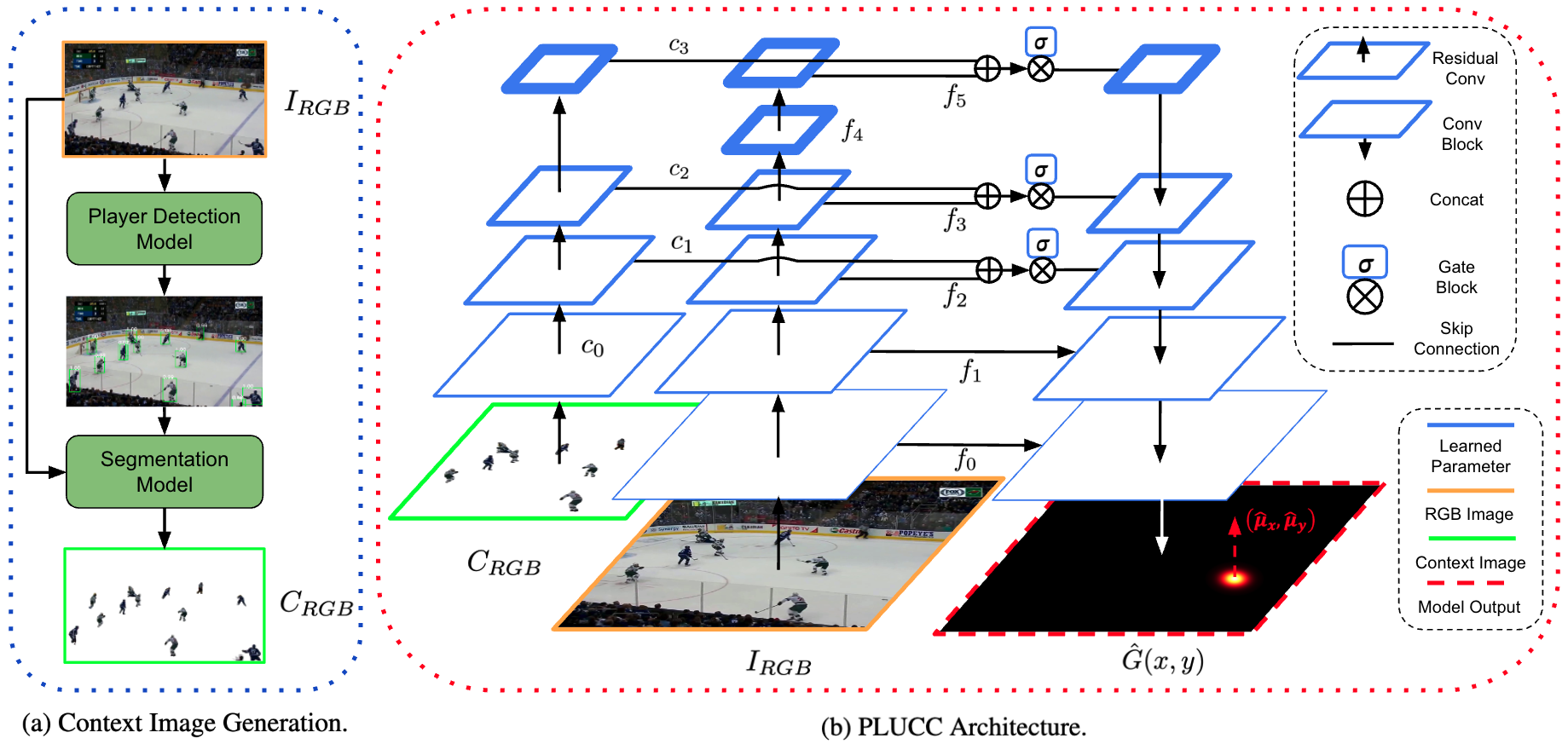}
    \caption{Left image (a) shows context image generation pipeline, a combination of detection and segmentation models frozen in training. $I_{RGB}$ is the full-resolution RGB frame, and $C_{RGB}$ is the half-resolution context image. Latent features are denoted as $f_n$ and $c_n$ for the feature pyramid encoder and context encoder, respectively. The right image (b) is the trained PLUCC model architecture, where $\hat{G}(x,y)$ is the predicted Gaussian heatmap.}
    \label{fig:model_arch}
    \vspace{-4mm}
\end{figure*}

The proposed PLUCC architecture, shown in Figure \ref{fig:model_arch}b, leverages a feature pyramid encoder and a context encoder that feed into the decoder through gated feature fusion, producing heatmaps. Context image ($C_{RGB}$) generation, shown in Figure \ref{fig:model_arch}a, uses a pre-trained player detection model followed by a segmentation model to create an RGB image consisting only of all the segmented players in that frame.

\subsubsection{Feature Pyramid Encoder}
The primary objective of the feature pyramid encoder is to capture the puck's features at various scales without strong contextual constraints. Following recent literature in small object detection \cite{survey_small_obj_wei}, the feature pyramid encoder processes full-resolution RGB images to capture the puck's features. The input image is defined as $I_{RGB} \in \mathbb{R}^{3 \times {H} \times W}$.  

The feature pyramid encoder is a modified ResNet-152 \cite{resnet} backbone taken from Wu et al. \cite{wuVoteCenter62022}, a choice that stems from its demonstrated capabilities in keypoint regression with high accuracies~\cite{wuVoteCenter62022}. 

We denote the feature pyramid encoder features as $\{f_0, f_1, f_2, f_3, f_4, f_5\}$, where \(f_0 \in \mathbb{R}^{B \times 64 \times H_0 \times W_0}\) is obtained after the initial convolution and \(f_5 \in \mathbb{R}^{B \times 1024 \times H_5 \times W_5}\) is the bottleneck feature.

\subsubsection{Context Encoder}
Relying solely on visual puck information is impractical for puck detection due to the small size of the puck and the dynamic nature of the game, which often involves significant occlusion and motion blur, as shown in Figures \ref{fig:dif_occluded} and \ref{fig:dif_blur}. Therefore, incorporating additional context is crucial to guide the model toward robust puck localization.

Prior research has revealed a significant correlation between the puck location and the position of the players in the rink \cite{vats2021puck, pidaparthy2019keep}. In Section \ref{sec:gaze}, we also highlight the correlation between player pose and the puck location. However, directly extracting player pose, as explored in \cite{balaji2024language}, is computationally expensive, and pose estimators often struggle under occlusion and motion blur.

To address these issues, we propose using an RGB segmentation mask of the players as the \textbf{contextual input}. Specifically, we first extract the bounding boxes of the players using a pretrained detector and then segment the detected players with a pretrained segmentation algorithm \cite{sam2}. Figure \ref{fig:model_arch}a highlights this process.

The {context image} is then defined as $C_{RGB} \in \mathbb{R}^{3 \times {\frac{H}{2}} \times {\frac{W}{2}}}$, where the $C_{RGB}$ is RGB segmentations of all the hockey players in $I_{RGB}$.

The context encoder processes the input context image to produce multi-scale features, denoted as $\{c_0, c_1, c_2, c_3\}$, where \(c_0 \in \mathbb{R}^{B \times 64 \times H_0 \times W_0}\) is obtained after the initial convolutional layer and is the only feature scale level that is not fused with the feature pyramid encoder.

To facilitate feature fusion in the decoder, the context encoder features are designed to match the same spatial scale of the deep feature pyramid encoder features such that:

\begin{itemize}
    \item \(c_1 \in \mathbb{R}^{B \times 256 \times H_2 \times W_2}\), matching the spatial resolution of \(f_2 \in \mathbb{R}^{B \times 512 \times H_2 \times W_2}\),
    \item \(c_2 \in \mathbb{R}^{B \times 512 \times H_3 \times W_3}\), matching the resolution of \(f_3 \in \mathbb{R}^{B \times 1024 \times H_3 \times W_3}\),
    \item \(c_3 \in \mathbb{R}^{B \times 1024 \times H_5 \times W_5}\), matching the resolution of \(f_5 \in \mathbb{R}^{B \times 1024 \times H_5 \times W_5}\).
\end{itemize}

Ensuring consistent dimensionality of multiscale features allows for seamless fusion in the decoder using convolutional layers.

\subsubsection{Gated Decoder}
At each decoding stage, features from the feature pyramid encoder and context encoder are concatenated and processed by a \emph{GateBlock} that learns static per-channel weights ($\gamma$). For an input tensor \(F \in \mathbb{R}^{B \times C \times H \times W}\), the GateBlock computes a gating vector
\begin{equation}
    g = \sigma(\gamma) \quad \text{with} \quad \gamma \in \mathbb{R}^{1 \times C \times 1 \times 1},
\end{equation}
where \(\sigma(\cdot)\) is the sigmoid function. The gated output is then given by

\begin{equation}
    \hat{F} = F \odot g,
\end{equation}
with \(\odot\) denoting element-wise multiplication.

The learned parameters \(\gamma\) remain static during inference, meaning the per-channel multiplication parameter is not dynamically changed based on the contents of input features fed to the gate. This static gating is reminiscent of Squeeze-and-Excitation blocks, which derive dynamic channel re-weightings based on global channel context \cite{squeeze} in inference.

When both the feature pyramid encoder feature \(f\) and the corresponding context encoder feature \(c\) are available, the fusion is defined as:
\begin{equation}
    \tilde{F} = \operatorname{ConvBlock}\Bigl( \operatorname{GateBlock}\Bigl( \operatorname{Up}(\tilde{F}_{\text{prev}}) \oplus f \oplus c \Bigr) \Bigr),
\end{equation}

\noindent where \(\operatorname{Up}(\cdot)\) upsamples the previous feature map \(\tilde{F}_{\text{prev}}\) to match the spatial dimensions of \(f\) and \(c\) using billinear interpolation, \(\oplus\) denotes concatenation, \(\operatorname{GateBlock}(\cdot)\) applies a sigmoid-based static gating operation, and \(\operatorname{ConvBlock}(\cdot)\) performs a 3\(\times\)3 convolution followed by batch normalization and ReLU activation.

In stages without a corresponding context feature (e.g., for \(f_0\) and \(f_1\)), the fusion is performed as:
\begin{equation}
    \tilde{F} = \operatorname{ConvBlock}\Bigl( \operatorname{Up}(\tilde{F}_{\text{prev}}) \oplus f \Bigr).
\end{equation}

The final heatmap logits are generated by upsampling the last refined feature map \(\tilde{F}_{\text{final}}\) to the original input resolution and applying a \(1 \times 1\) convolution:
\begin{equation}
    \tilde{Z}(x,y) = \operatorname{Conv}_{1\times1}\Bigl( \operatorname{Up}(\tilde{F}_{\text{final}}) \Bigr).
\end{equation}
where $\tilde{Z}(x,y)$ represents the raw heatmap logits. 

Lastly, the final predicted Gaussian heatmap $\hat{G}(x,y)$ is derived from normalizing and using a softmax operation over the logits spatial domain:
\begin{equation}
    \hat{G}(x,y) = \frac{\exp(\tilde{Z}(x,y))}{\sum_{x',y'} \exp(\tilde{Z}(x',y'))}.
\end{equation}

\subsection{Objective Function}
The PLUCC model is trained to minimize the distance between the predicted $\hat{G}(x,y)$ and ground truth $G(x,y)$ Gaussian heatmaps. To do so, we utilize Kullback-Leibler divergence loss ($\mathcal{L}_{KL}$) to penalize spatial displacement, encouraging the network to predict a peak precisely at the correct point. 

To measure the similarity between the predicted and ground truth distributions, the KL divergence loss is computed as:
\begin{equation}
    \mathcal{L}_{KL} = \frac{1}{B} \sum_{i=1}^{B} \sum_{x,y} G_i(x,y) \log \frac{G_i(x,y)}{\hat{G}_i(x,y)},
\end{equation}
where $B$ is the batch size, and the loss is averaged across all samples in the batch to ensure stability.

\noindent \textbf{Gaussian Heatmap Label.}
Gaussian heatmaps provide a soft training target with peak confidence at the object center. Inspiration for heatmap training stems from Tarashima et al.'s \cite{tarashima_widely_2023} baseline for sports ball detection, where they achieve enhanced accuracy by employing position-aware model training with ground truth heatmaps, especially for small, fast-moving balls. This soft representation explicitly models spatial uncertainty and directs the model’s attention precisely to the object center.

To derive a Gaussian heatmap label for training, the center of the ground truth bounding boxes \((\mu_x, \mu_y)\) is the peak of the two-dimensional Gaussian. For each pixel \((x,y)\) in the image, the Gaussian value is computed as:

\begin{equation}
    G_{gt}(x, y) = \exp\left(-\frac{(x - \mu_x)^2 + (y - \mu_y)^2}{2\sigma^2}\right)
\end{equation}
where $\sigma$ is the variance of the Gaussian label, a parameter that determines how soft or granular our target label is.

%% file: sec/4_experiments.tex
\section{Experimentation}

\subsection{Datasets}
The VIP-PuckDataset is an expansive proprietary dataset comprising 150,000 images of annotated puck bounding boxes and segmentations sorted into categories of challenges, including standard, blurry, yellow-line, and goal-line pucks, shown in Figure \ref{fig:dif_blur}, \ref{fig:dif_yellow}, \ref{fig:dif_goal} respectively. The training, validation, and test splits were created such that no frames from a single match belong to different dataset splits, thus ensuring our models have truly generalized to puck detection and not overfit a singular game dynamics. 115,000 frames were used in training, and 16,000 for validation. The test set, comprised of 19,000 frames, only contains challenging scenarios, such as blurry (\ref{fig:dif_blur}), goal-line (\ref{fig:dif_goal}), and yellow-line (\ref{fig:dif_yellow}) pucks, to evaluate the model’s performance under difficult conditions. 

\subsection{Implementation Details}
The PLUCC model consists of 114 million trainable parameters and was trained on an NVIDIA H100 over 25 epochs and with a batch size of 10. At inference, PLUCC and context image generation operate together at 6.03 frames per second. The feature pyramid encoder input ($I_{RGB}$) size was 720 by 1280 (height by width) images, and the context encoder was fed 360 by 640 context images ($C_{RGB}$). The initial learning rate was 0.001 and would be reduced by a factor of ten if no improvement over the validation set occurred for five epochs. Lastly, a Gaussian variance ($\sigma$) of 5.0 was used, with further reasoning highlighted in Section \ref{sec:sigma_param}. Training data augmentation included random flipping, Gaussian blurring, added noise, and normalization.

\noindent \textbf{Context-driven dropout.} To prevent the PLUCC model from becoming over-reliant on direct visual cues from the feature pyramid encoder, $I_{RGB}$ images are withheld during training with a probability of $p_{drop} = 1\%$. This forces the network to regress the puck location solely from contextual cues such as implicit player pose and position formations of the players. This mechanism is particularly important in scenarios where the puck is completely occluded or visually indistinguishable, with its effects further explored in Section \ref{sec:cd_dropout}.



\subsection{Baselines}
Our comparison for model performance compares the two most commonly used open-source models, Faster-RCNN \cite{ren_faster_2016} and YOLOv5 \cite{yolo}, trained on our VIP-PuckDataset for a baseline comparison \cite{yang2021puck, sarkhoosh_ai-based_2024}. Furthermore, we trained a modified Fully-Convolutional-ResNet152 \cite{wuVoteCenter62022} (FCN-ResNet152) as another baseline model for producing heatmap outputs. The FCN-ResNet152 is architecturally similar to the PLUCC model, forgoing the context encoder and GateBlocks. Thus, the FCN-ResNet152 provides a direct evaluation of the proposed model improvement when leveraging the context encoder and the gating. 

\begin{figure*}[t]
    \centering
    \begin{subfigure}{\columnwidth}
        \centering
        \includegraphics[width=\linewidth]{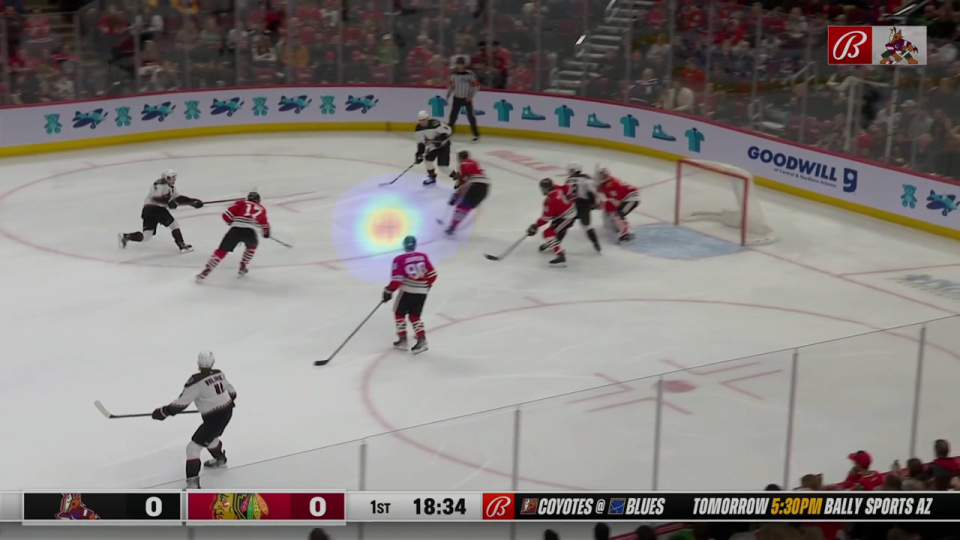}
        \caption{Blur}
        \label{fig:sub1}
    \end{subfigure}
    \hfill
    \begin{subfigure}{\columnwidth}
        \centering
        \includegraphics[width=\linewidth]{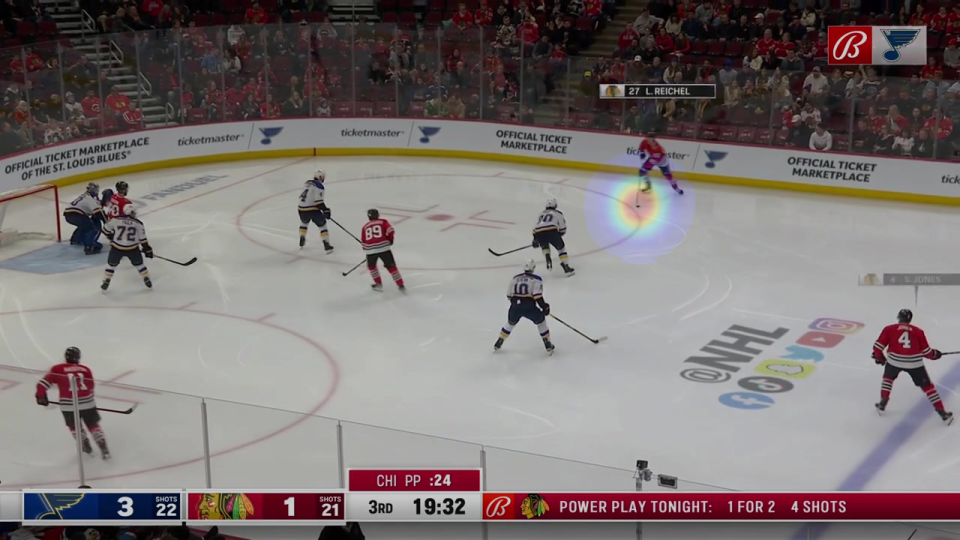}
        \caption{Artifact}
        \label{fig:sub2}
    \end{subfigure}
    
    \vspace{1.0em} 

    \begin{subfigure}{\columnwidth}
        \centering
        \includegraphics[width=\linewidth]{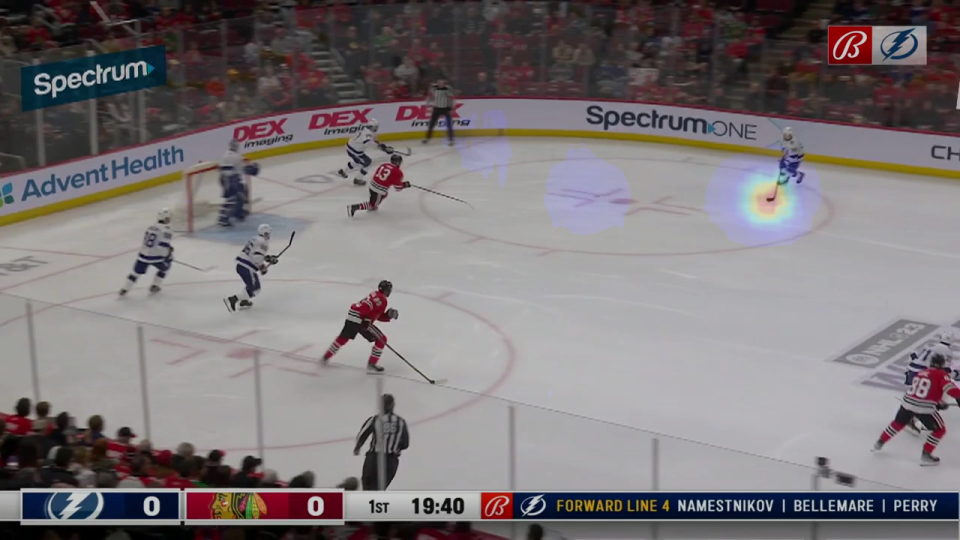}
        \caption{Occluded}
        \label{fig:sub3}
    \end{subfigure}
    \hfill
    \begin{subfigure}{\columnwidth}
        \centering
        \includegraphics[width=\linewidth]{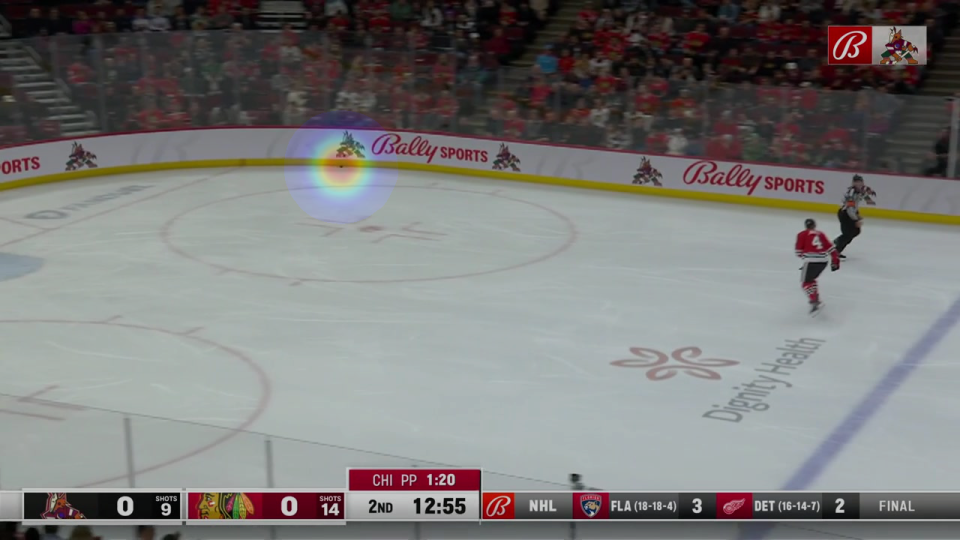}
        \caption{Yellow-line}
        \label{fig:sub4}
    \end{subfigure}
    \vspace{-3mm}
   \caption{\textbf{Qualitative results of PLUCC with expanded Gaussian overlays}. Sub-figure (a) demonstrates robust detection under heavy puck blurring conditions, (b) shows model resistance to out-of-distribution broadcast artifacts, (c) highlights a correct puck detection even under full occlusion from a hockey stick, and (d) showcases the model's ability to predict the puck location when it is on a yellow-line, where the contrast between the puck and the background differs significantly from commonly occurring scenarios.}
    \vspace{-1mm}
    \label{fig:results}
\end{figure*}

\subsection{Evaluation Metrics}
\subsubsection{Image Coordinate-Space Localization Error}
To ensure a fair comparison across detection algorithms that produce heatmaps and those that produce bounding boxes, we calculate the pixel Euclidean distance ($\mathcal{D}_{pixel}$) between the predicted puck center ($\hat{\mu}_{x},\hat{\mu}_{y}$) and the ground truth center ($\mu_{x},\mu_{y}$). 
\begin{equation}
    \mathcal{D}_{pixel} = \sqrt{(\hat{\mu}_{x} - \mu_{x})^2 + (\hat{\mu}_{y} - \mu_{y})^2}
\end{equation}

The puck location is derived from a predicted heatmap (\(\hat{G}(x,y)\))
by taking the maximum point:
\begin{equation}
    \hat{\mu}_x, \hat{\mu}_y = \arg\max_{(x,y)} \hat{G}(x,y).
\end{equation}

Bounding box puck centers are taken from the center-most point of the bounding box such that:
\begin{equation}
\mu_x, \mu_y = \left(\frac{\hat{x}_{\text{min}} + \hat{x}_{\text{max}}}{2}, \frac{\hat{y}_{\text{min}} + \hat{y}_{\text{max}}}{2} \right).
\end{equation}

Furthermore, we do not compute $\mathcal{D}_{pixel}$ for test samples that do not have a ground truth label. Models like YOLOv5 \cite{yolo} and Faster-RCNN \cite{ren_faster_2016} can detect the object when it appears in the frame. Thus, our evaluation adds to the average $\mathcal{D}_{pixel}$ metric for only visible puck labels. This ensures that our comparison holds true across detection approaches, whether they explicitly predict object presence or rely solely on heatmap-based localization.

Average Precision is computed with a distance threshold \(\tau \in \{5, 10, 25, 50\}\) pixels. A prediction is considered correct if the Euclidean distance $\mathcal{D}_{pixel}$ between the predicted puck center and the ground truth puck center is within the threshold $\tau$. The mean Average Precision ($mAP^{\tau}$) is computed as the mean of $AP^\tau$ values across all thresholds, providing a holistic measure of model performance across varying levels of localization accuracy.

\subsubsection{Rink-Space Localization Error}
\noindent The Rink-Space Localization Error (RSLE) compensates for the perspective distortion present in broadcast footage, where the same pixel error can correspond to vastly different real-world distances depending on the puck’s location relative to the camera. Specifically, a pixel error ($\mathcal{D}_{pixel}$) near the camera, due to the viewing angle, may represent less of a physical distance on the rink compared to an error of $\mathcal{D}_{pixel}$ near the far boards. By transforming image coordinates into a standardized rink space, RSLE ensures that localization errors are measured in consistent physical units, removing the inherent bias of comparisons in image coordinate-space. 

The RSLE metric for a single puck comparison can be formulated as:


\begin{table}[t]
\begin{center}
\caption{\textbf{Puck detection performance in image coordinates.} Metrics reported are mean Average Precision ($mAP^{\tau}$), Average Precision ($AP^{\tau}$ for $\tau \in \{5,10,25,50\}$), average Euclidean error ($\mathcal{D}_{pixel}$), and inference speed (FPS) on RTX6000. FPS marked by $^*$ includes preprocessing time.}
\label{tab:image_coord}
\vspace{-3mm}
\resizebox{1\columnwidth}{!}{
\begin{tabular}{l|ccccccc}
\toprule
Method & $mAP^{\tau}$ & $AP^{5}$ & $AP^{10}$ & $AP^{25}$ & $AP^{50}$ & $\mathcal{D}_{pixel}$ & FPS\\ 
\midrule    
YOLOv5-Large \cite{yolo} & 56.3 & 51.5 & 57.5 & 57.9 & 58.4 & 55.2 & \textbf{128.9} \\
Faster-RCNN \cite{ren_faster_2016} & 71.3 & 69.6 & 70.9 & 72.0 & 73.0 & 71.4 & 50.1 \\
\midrule
FCN-ResNet152 \cite{wuVoteCenter62022} & 79.6 & 74.8 & 79.9 & 81.3 & 82.4 & 52.11 & 54.1 \\
($\sigma = 5$) & & & & & & & \\
\midrule
\textbf{PLUCC ($\sigma = 15$)} & 81.6 & 76.2 & 82.0 & 83.7 & 84.7 & 48.8 & 53.3 (6.0)$^*$\\
\textbf{PLUCC ($\sigma = 5$)} & \textbf{83.5} & \textbf{82.2} & \textbf{83.6} & \textbf{84.2} & \textbf{85.0} & \textbf{47.0} & 53.3 (6.0)$^*$\\
\bottomrule
\end{tabular}
}
\end{center}
\vspace{-4mm}
\end{table}


\begin{equation}
    RSLE = \sqrt{(\hat{x}_{rink} - x_{rink})^2 + (\hat{y}_{rink} - y_{rink})^2}
\end{equation}

\noindent where, $\hat{x}_{rink}$, $\hat{y}_{rink}$, $x_{rink}$ and $x_{rink}$ are the transformed estimated and ground truth puck location in homography coordinates respectively.

Rink Space Localization Error Average Precision ($AP^r$) is the percentage of detections transformed to rink coordinates that lie within the puck radius, where $r=3.81 cm$ is the puck radius~\cite{puck_size}. $AP^{r\times2}$ is the percentage of detection within the puck diameter (twice the radius), and $AP^{r\times4}$ is the percentage within twice the diameter. 

\noindent \textbf{Homography Estimation.} In our approach, we use the homography matrix $\mathbf{H}$ to map puck positions from the image plane to rink coordinates. This matrix establishes a correspondence between the 2D image coordinates and a predefined homographic space and is computed for each frame using the technique proposed by Shang et al. \cite{homography}.

The transformation of the puck's image coordinates $(\mu_x,\mu_y)$ into rink-space is performed in two steps. First, we warp the coordinates into a top-down view of the rink segmentation (with dimensions 720 by 1280), and then we scale these warped positions to match the rink dimensions.

First, the puck's position is represented in homogeneous coordinates:
\begin{equation}
    \tilde{\mathbf{p}} = \begin{bmatrix} \mu_x \\ \mu_y \\ 1 \end{bmatrix} = \begin{bmatrix} \tilde{p}_x \\ \tilde{p}_y \\ \tilde{p}_z \end{bmatrix}.
\end{equation}

Next, we apply the homography matrix to transform this point:
\begin{equation}
    \tilde{\mathbf{p}}' = \mathbf{H} \, \tilde{\mathbf{p}}.
\end{equation}

To convert the result back to inhomogeneous coordinates, we normalize by the third component:
\begin{equation}
    (p_{\text{warp},x}, p_{\text{warp},y}) = \left( \frac{\tilde{p}'_x}{\tilde{p}'_z}, \, \frac{\tilde{p}'_y}{\tilde{p}'_z} \right).
\end{equation}

Finally, using the dimensions of the homographic template (720 by 1280) and the standard NHL rink size (length $L=61$ m and width $W=25.9$ m) \cite{athletica_size_of_hockey_rinks}, we scale the warped coordinates to obtain the final rink-space coordinates:
\begin{equation}
    x_{\text{rink}} = \frac{p_{\text{warp},x}}{1280} \times L, \quad y_{\text{rink}} = \frac{p_{\text{warp},y}}{720} \times W.
\end{equation}

\subsection{Main Results}

The results in Table \ref{tab:image_coord} compare our PLUCC method with other single-frame puck detection baseline models \cite{ren_faster_2016, yolo} and the FCN-ResNet152 \cite{wuVoteCenter62022}, evaluated in image coordinates on the VIP-PuckDataset test set. PLUCC outperforms baseline object detection models across all values of $\tau$, demonstrating a 12.3\% increase in $mAP^\tau$ over Faster-RCNN, the best-performing baseline. Furthermore, the observed 3.9\% improvement in $mAP^\tau$ compared to the FCN-ResNet152 underscores the effectiveness of incorporating the context encoder and gated decoder into our model architecture. Although the inference speed of our model is significantly impacted by the preprocessing required to generate $C_{RGB}$, the standalone inference speed (without preprocessing) remains comparable to FCN-ResNet152. Since preprocessing steps like player detection and segmentation are integral to hockey analytics tasks such as player tracking \cite{prakash2024multi} and pose estimation \cite{vasilikopoulos2024dposedepthintermediaterepresentation}, the performance of PLUCC aligns closely with existing methods. Figure \ref{fig:results} illustrates the Gaussian heatmap outputs of our model, demonstrating robust detections under challenging conditions.




\begin{table}[t]
\begin{center}
\caption{\textbf{Comparison of comparable models in homographic rink-space coordinates.} ${AP}^{r}$ represents the average precision when the predicted puck location falls within the puck radius ($r=3.81$ cm), ${AP}^{r\times2}$ measures average precision when the prediction is within twice the puck radius (puck diameter), and ${AP}^{r\times4}$ measures average precision of predictions within four times the radius (twice the diameter). The average rink-space localization error is the $RSLE_{avg}$ metric in meters.}
\label{tab:homographic}
\vspace{-3mm}
\resizebox{\columnwidth}{!}{
\begin{tabular}{l|cccc}
\toprule
Method & ${AP}^{r}$ & ${AP}^{r\times2}$ & ${AP}^{r\times4}$ & $RSLE_{avg}\space(m)$ \\ 
\midrule
YOLOv5-Large \cite{yolo} & 0.17 & 2.69 & 43.11 & 3.14 \\ 
Faster-RCNN \cite{ren_faster_2016} & 18.58 & 19.14 & 20.06 & 3.76 \\
\midrule
FCN-ResNet152($\sigma = 5$)\cite{wuVoteCenter62022} & 41.85 & 54.23 & 62.74 & 3.89 \\
\midrule
\textbf{PLUCC ($\sigma = 15$)}& 20.69 & 40.35 & 67.68 & 1.10 \\
\textbf{PLUCC ($\sigma = 5$)}& \textbf{43.59} & \textbf{62.94} & \textbf{81.23} & \textbf{1.05} \\
\bottomrule
\end{tabular}
}
\end{center}
\vspace{-4mm}
\end{table}

Results shown in Table \ref{tab:homographic} highlight the PLUCC method's superior detection accuracy in the rink-space coordinates, surpassing the strongest baseline object detector, Faster-RCNN \cite{ren_faster_2016} by 25.01\% in $mAP^r$, 43.8\% in $mAP^{r\times{2}}$, and is on average 2.71 meters closer to the puck. Furthermore, the model outperforms the FCN-ResNet152 \cite{wuVoteCenter62022} by 1.74\% in $mAP^r$, demonstrating the strength of the refinement the context encoding provides.

\begin{figure}[t]
    \centering
    \begin{subfigure}{0.23\textwidth}
        \centering
        \includegraphics[width=\linewidth]{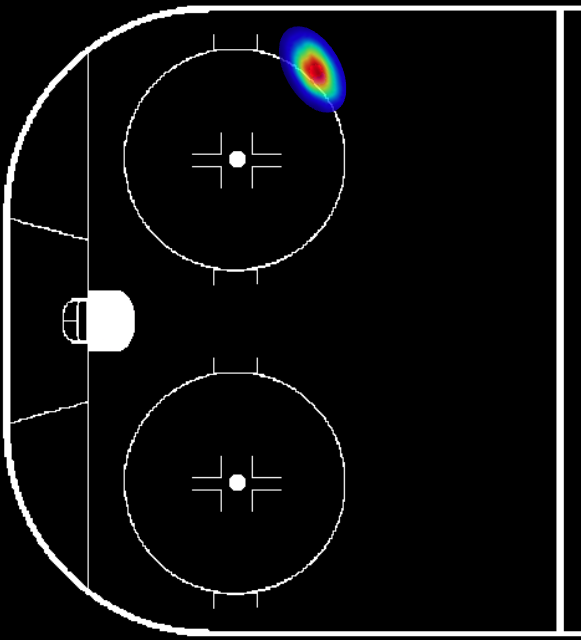}
        \caption{Puck heatmap in homography.}
        \label{fig:sub1}
    \end{subfigure}
    \hfill
    \begin{subfigure}{0.23\textwidth}
        \centering
        \includegraphics[width=\linewidth]{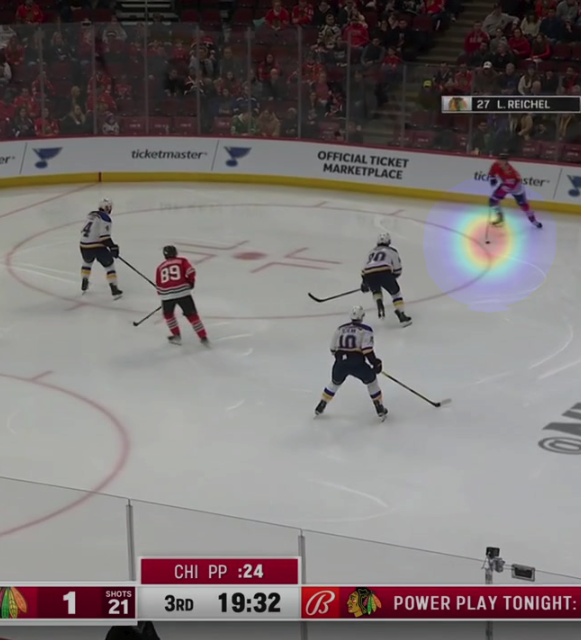}
        \caption{Puck heatmap.}
        \label{fig:sub2}
    \end{subfigure}
    \vspace{-2mm}
    \caption{Visualization of (a) heatmap detection transformed to homographic coordinates, and (b) puck detection heatmap overlayed on the processed frame.}
    \label{fig:results_homography}
\end{figure}

\subsection{Ablation}
Our ablation studies highlight the performance of an independent context encoder, the selection of the optimal Gaussian variance ($\sigma$) parameter, and the effects of including context-driven dropout in training.  
\subsubsection{Independent Context Encoder}
\label{sec:gaze}
\begin{figure}[t]
    \centering
    \includegraphics[width=0.8\linewidth]{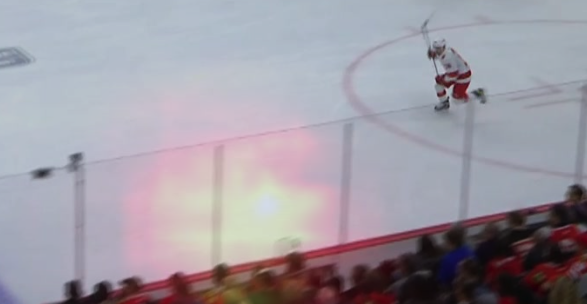}
    \caption{Heat map generated by only the context encoder when processing a single segmented player. The network, trained independently on context images, highlights regions in the player's line of sight, implicitly predicting the occluded puck's location behind the boards.}
    \label{fig:context-only}
    \vspace{-4mm}
\end{figure}

Training the FCN-ResNet152 \cite{wuVoteCenter62022} independently on context images $C_{RGB}$ (no raw RGB images including puck features) resulted in poor detection accuracies, highlighted in Table \ref{tab:context_only_results}. However, this ablation experiment produced heat maps that peak in regions corresponding to the direction of the player's gaze, shown in Figure \ref{fig:context-only}. This implies that the context encoder learns to look where players focus, an important cue in hockey where a player's line of sight often correlates with puck location. 

\begin{table}[t]
\begin{center}
\caption{Performance comparison of the PLUCC model and an FCN-ResNet152 trained independently on context images.}
\label{tab:context_only_results}
\vspace{-3mm}
\resizebox{1\columnwidth}{!}{
\begin{tabular}{l|cccccc}
\toprule
Method & $mAP^{\tau}$ & $AP^{5}$ & $AP^{10}$ & $AP^{25}$ & $AP^{50}$ & $\mathcal{D}_{pixel}$\\ 
\midrule
FCN-ResNet152 & 6.0 & 0.2 & 1.0 & 5.7 & 18.0 & 178.5 \\
(Only Context Image) & & & & & & \\
\midrule
\textbf{PLUCC ($\sigma = 5$)} & \textbf{83.5} & \textbf{82.2} & \textbf{83.6} & \textbf{84.2} & \textbf{85.0} & \textbf{47.0} \\
\bottomrule
\end{tabular}
}
\end{center}
\vspace{-4mm}
\end{table}

\subsubsection{Sigma Parameter Selection}
\label{sec:sigma_param}

\begin{figure}[t]
    \centering
    \includegraphics[width=0.9\linewidth]{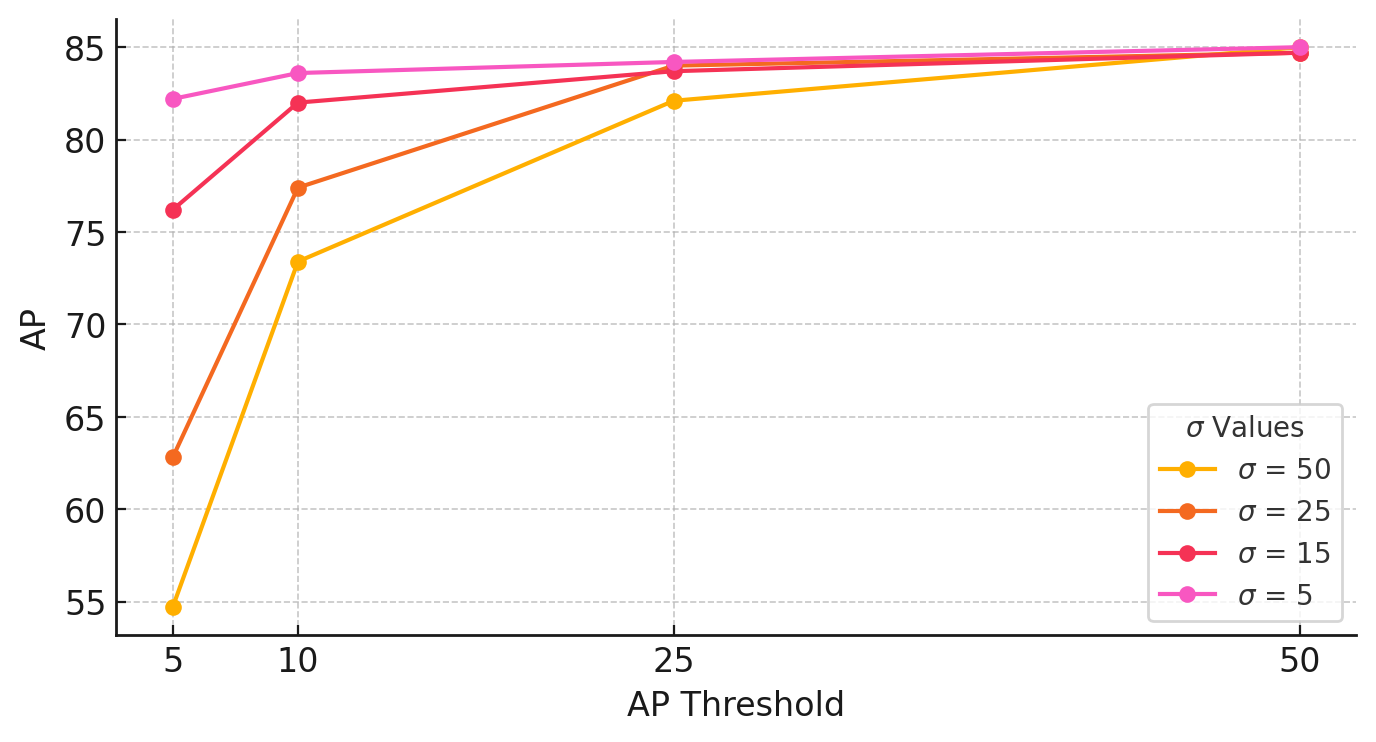}
    \caption{Comparison of performance of different models trained with different target Gaussian heatmap variances ($\sigma$).}
    \label{fig:ap_graph}
\end{figure}

To select the optimal Gaussian heatmap variance, we conducted an ablation study by training models with different $\sigma$ values. Figure \ref{fig:ap_graph} highlights the varying model's performances over average precision threshold $\tau \in {5,10,25,50}$ on the test set, where the model trained with $\sigma=5$ achieved the best performance across all thresholds $\tau$. The model performs best at lower thresholds (e.g., $\tau=5$ and $\tau=10$), indicating that a smaller $\sigma$ results in a more focused heatmap. Furthermore, at higher thresholds (e.g., $\tau=25$ and $\tau=50$), the performance of all models converges, suggesting that the benefit of a more granular heatmap is most significant when exact localization accuracy is required. 

\subsubsection{Context-driven Dropout}
\label{sec:cd_dropout}
Table \ref{tab:context_dropout} highlights the performance boost of including a dropout of the $I_{RGB}$ input image into the feature pyramid encoder with a probability of $p_{drop} = 1\%$, the network is compelled to leverage information from the contextual image ($C_{RGB}$) — such as player positions and orientations — which are indicative of the puck's location. The increase in accuracy highlights the necessity of forcing the model to rely more on contextual cues when the direct visual features are unreliable. 

\begin{table}[t]
\begin{center}
\caption{Comparison between PLUCC network trained with context-driven dropout probability of 1\%, and PLUCC trained with no dropout.}
\label{tab:context_dropout}
\vspace{-3mm}
\resizebox{\columnwidth}{!}{
\begin{tabular}{l|cc|cc}
\toprule
Method & ${mAP}^{5}$& $\mathcal{D}_{pixel}$ & ${AP}^{r}$ & $RSLE_{avg}(m)$   \\ 
\midrule
\textbf{PLUCC ($p_{drop} = 0\%$)}& 78.7 & 61.15 & 39.9 & 1.36 \\
\textbf{PLUCC ($p_{drop} = 1\%$)}& \textbf{82.2} & \textbf{47.0} & \textbf{43.6} & \textbf{1.05} \\
\bottomrule
\end{tabular}
}
\end{center}
\vspace{-4mm}
\end{table}

%% file: sec/5_conclusions.tex
\section{Conclusion}

This paper introduces PLUCC, a novel, context-aware detection framework that fuses multi-scale visual features with player cues to localize pucks in ice hockey broadcast videos. Our method tackles challenges such as motion blur, occlusions, and perspective distortion, leveraging a dedicated feature pyramid encoder, context encoder, and gated decoder. It is evaluated using the novel RSLE metric to offer a fair, homography-based evaluation of detection accuracy. Experiments on the PuckDataset show that PLUCC outperforms state-of-the-art baselines by boosting average precision by over 12\% in image and 25\% in rink coordinates. 

These findings pave the way for highly robust tracking systems, akin to Tarashima et al.'s \cite{tarashima_widely_2023} accurate tracking with heatmaps over time. Future work could further increase accuracy by integrating advanced channel fusion methods like squeeze-and-excitation \cite{squeeze}, channel attention, or transformers. Other sports, such as lacrosse, have prolonged occlusions; thus, applying PLUCC could be practical in such domains. 

The improvements in detection accuracy could transform how coaches, teams, and broadcasters analyze game dynamics, leading to enhanced strategic decisions and more engaging fan experiences \cite{kinexon2025}. Moreover, by offering a robust, cost-effective alternative to expensive tracking systems like Hawk-Eye \cite{hawkeye, hawkeye_2}, PLUCC is a strong alternative for smaller organizations such as amateur leagues.


%% file: sec/acknowledgments.tex
\section*{Acknowledgments}
This work was supported by a grant with the Natural Sciences and Engineering Research Council (NSERC) partnered with Stathletes, Inc. Stathletes also provided the puck annotation data used in this research.
